\documentclass[11pt,a4paper]{article}
\usepackage{times,latexsym}
\usepackage{url}
\usepackage[T1]{fontenc}

\usepackage[acceptedWithA]{tacl2021v1}

\usepackage{latexsym}
\usepackage{mathptmx} 

\usepackage{hyperref}
\usepackage{xcolor}

\usepackage{graphicx,epstopdf}			

\graphicspath{{graphics/}} 

\usepackage{natbib}

\usepackage{times}
\usepackage{latexsym}
\usepackage{url}

\usepackage{lipsum} 
\usepackage{stmaryrd}
\usepackage{bbm}

\usepackage{amsmath}
\usepackage{amsthm}
\usepackage{amsfonts}
\usepackage{bm}
\usepackage{thmtools,mathtools}
\usepackage{thm-restate}
\usepackage[T1]{fontenc}
\usepackage{booktabs}
\usepackage{multirow}
\usepackage{booktabs}
\usepackage{verbatim}
\usepackage{mdwlist}
\usepackage[ruled,norelsize,linesnumbered]{algorithm2e}
\usepackage{dsfont}
\usepackage[shortlabels]{enumitem}
\usepackage{subcaption}
\usepackage{array}
\usepackage{pifont}
\usepackage{CJKutf8}

\usepackage{multicol}
\usepackage{float}
\usepackage{lipsum}
\usepackage[font=small]{caption}

\usepackage{xspace}

\newcommand{\mt}{\textsc{mt}\xspace}
\newcommand{\nmt}{\textsc{nmt}\xspace}

\newcommand{\bleu}{\textsc{bleu}\xspace}

\newcommand{\wmt}{\textsc{wmt}\xspace}

\newcommand{\lrp}{\textsc{lrp}\xspace}
\newcommand{\xlmr}{\textsc{xlm-r}\xspace}
\newcommand{\laser}{\textsc{laser}\xspace}
\newcommand{\cometqe}{\textsc{comet-qe}\xspace}
\newcommand{\comet}{\textsc{comet}\xspace}

\newcommand{\LowSrcContribHypo}{Low Source Contribution Hypothesis\xspace}
\newcommand{\LocalSrcContribHypo}{Local Source Contribution Hypothesis\xspace}
\newcommand{\StaticSrcContribHypo}{Static Source Contribution Hypothesis\xspace}
\newcommand{\NormSrcContrib}{Normalized Source Contribution\xspace}
\newcommand{\HighContribRatio}{High-Contribution Ratio\xspace}
\newcommand{\SrcContribStaticity}{Source Contribution Staticity\xspace}

\newcommand{\mlp}{\textsc{mlp}\xspace}

\theoremstyle{definition}

\makeatletter
\newcommand{\removelatexerror}{\let\@latex@error\@gobble}
\makeatother

\usepackage{enumitem}
\setlist{nolistsep,leftmargin=*}

\title{Understanding and Detecting Hallucinations in \\ Neural Machine Translation via Model Introspection}

\author{Weijia Xu \\
	University of Maryland \\
	{\tt \href{mailto:weijia@cs.umd.edu}{weijia@cs.umd.edu}} \\\And
        Sweta Agrawal \\
        University of Maryland \\
	{\tt \href{mailto:sweagraw@cs.umd.edu}{sweagraw@cs.umd.edu}} \\\And
        Eleftheria Briakou \\
        University of Maryland \\
	{\tt \href{mailto:ebriakou@cs.umd.edu}{ebriakou@cs.umd.edu}} \\\AND
        Marianna J. Martindale \\
        University of Maryland \\
	{\tt \href{mailto:mmartind@umd.edu}{mmartind@umd.edu}} \\\And
	Marine Carpuat \\
	University of Maryland \\
	{\tt \href{mailto:marine@cs.umd.edu}{marine@cs.umd.edu}} \\}

\begin{document}
\maketitle
\begin{abstract}
    Neural sequence generation models are known to ``hallucinate'', by producing outputs that are unrelated to the source text. These hallucinations are potentially harmful, yet it remains unclear in what conditions they arise and how to mitigate their impact. In this work, we first identify internal model symptoms of hallucinations by analyzing the relative token contributions to the generation in contrastive hallucinated vs. non-hallucinated outputs generated via source perturbations.  We then show that these symptoms are reliable indicators of natural hallucinations, by using them to design a lightweight hallucination detector which outperforms both model-free baselines and strong classifiers based on quality estimation or large pre-trained models on manually annotated English-Chinese and German-English translation test beds. 
\end{abstract}

\section{Introduction}

\begin{table*}[t!]
\centering
\scalebox{0.92}{
\begin{tabular}{p{2cm}p{12.8cm}}
\toprule
\multicolumn{2}{l}{\textbf{Counterfactual hallucination from perturbation}} \\
\textit{Source} &  Republicans Abroad are not running a similar election, nor will they have delegates at the convention. Recent elections have emphasized the value of each vote.\\
\textit{Good \nmt} &  \begin{CJK*}{UTF8}{gbsn}国外的共和党没有举行类似的选举，也没有代表参加大会。最近的选举强调了每次投票的价值。\end{CJK*} \\
\textit{Perturbed Source} & Repulicans Abroad ar not runing a simila election, nor will they have delegates at the convention. Recent elections have emphasized the value o each vote.\\
\textit{Hallucination} & \begin{CJK*}{UTF8}{gbsn}大耳朵评论管理人员有权保留或删除其管辖评论中的任意内容。 \end{CJK*}\\
& \textit{Gloss: The big ear comments that administrators have the right to retain or delete any content in the comments under their jurisdiction.}\\
\midrule
\multicolumn{2}{l}{\textbf{Natural hallucination}} \\
\textit{Source} & DAS GRUNDRECHT JEDES EINZELNEN AUF FREIE WAHL DES BERUFS, DER AUSBILDUNGSSTÄTTE SOWIE DES AUSBILDUNGS - UND BESCHÄFTIGUNGSORTS MUSS GEWAHRT BLEIBEN.\\
& \textit{Gloss: The fundamental right of every individual to freely choose their profession, their training institution and their employment place must remain guaranteed.} \\
\textit{Hallucination} & THE PRIVACY OF ANY OTHER CLAIM, EXTRAINING STANDARDS, EXTRAINING OR EMPLOYMENT OR EMPLOYMENT WILL BE LIABLE.\\
\bottomrule
\end{tabular}
}
\caption{Contrasting counterfactual English-Chinese hallucinations derived from source perturbations (top) with a natural hallucination produced by a German-English \nmt model (bottom).}
\label{tab:example}
\vspace{-10pt}
\end{table*}

\looseness=-1
While neural language generation models can generate high quality text in many settings, they also fail in counter-intuitive ways, for instance by  ``hallucinating''~\cite{WisemanSR2017,LeeFiratAgarwalFannjiangSussillo2018,Falke2019}. In the most severe case, known as  ``detached hallucinations'' \cite{RaunakMenezesJunczys-Dowmunt2021}, the output is completely detached from the source, which not only reveals fundamental limitations of current models, but also risks misleading users and undermining trust~\cite{BenderGMS2021,MartindaleCarpuat2018}. 
Yet, we lack a systematic understanding of the conditions where hallucinations arise, as hallucinations occur infrequently among translations of naturally occurring text. As a workaround, prior work has largely focused on black-box detection methods which train neural classifiers on synthetic data constructed by heuristics~\cite{Falke2019,ZhouNGDGZG2020}, and on studying hallucinations given artificially perturbed inputs~\cite{LeeFiratAgarwalFannjiangSussillo2018,Shi2022}. 

\looseness=-1
In this paper, we address the problem by first identifying the internal model symptoms that characterize hallucinations given artificial inputs and then testing the discovered symptoms on translations of natural texts.
Specifically, we study hallucinations in Neural Machine Translation (\nmt) using two types of interpretability techniques: saliency analysis and perturbations. We use saliency analysis \cite{Bach2015,Voita2021} to compare the relative contributions of various tokens to the hallucinated vs. non-hallucinated outputs generated by diverse adversarial perturbations in the inputs~(Table~\ref{tab:example}) inspired by~\citet{LeeFiratAgarwalFannjiangSussillo2018,RaunakMenezesJunczys-Dowmunt2021}.
Results surprisingly show that source contribution patterns are stronger indicators of hallucinations than the relative contributions of the source and target, as had been previously hypothesized~\cite{Voita2021}. We discover two distinctive source contribution patterns, including~1) concentrated contribution from a small subset of source tokens, and~2) the staticity of the source contribution distribution along the generation steps~(\S~\ref{sec:exp_perturbed_hallucinations}). 

We further show that the symptoms identified generalize to hallucinations on natural inputs by using them to design a lightweight hallucination classifier (\S~\ref{sec:detection}) that we evaluate on manually annotated hallucinations from English-Chinese and German-English \nmt (Table~\ref{tab:example}). Our study shows that our introspection-based detection model largely outperforms model-free baselines and the classifier based on quality estimation scores. Furthermore, it is more accurate and robust to domain shift than black-box detectors based on large pre-trained models (\S~\ref{sec:exp_natural_hallucinations}).

Before presenting these two studies, we review current findings about the conditions in which hallucinations arise and formulate three hypotheses capturing potential hallucination symptoms.

\section{Hallucinations: Definition and Hypotheses}
\label{sec:hypotheses}

The term ``hallucinations'' has varying definitions in \mt and natural language generation. We adopt the most widely used one, which refers to output text that is unfaithful to the input~\cite{MaynezNBM2020,ZhouNGDGZG2020,XiaoW2021,Ji2022Survey}, while 
others include fluency criteria as part of the definition \citep{WangSennrich2020,MartindaleCarpuatDuhMcNamee2019a}.
Different from the previous work that aims to detect partial hallucinations at the token level~\cite{ZhouNGDGZG2020}, we focus on \textbf{detached hallucinations} where a major part of the output is unfaithful to the input, as these represent severe errors, as illustrated in Table~\ref{tab:example}.

Prior work on understanding the conditions that lead to hallucinations has focused on training conditions and data noise~\cite{Ji2022Survey}. For \mt, \citet{RaunakMenezesJunczys-Dowmunt2021} show that hallucinations under perturbed inputs are caused by training samples in the long tail that tend to be memorized by Transformer models, while natural hallucinations given unperturbed inputs can be linked to corpus-level noise. \citet{BriakouC2021} show that  models trained on samples where the source and target side diverge semantically output degenerated text more frequently. \citet{WangSennrich2020} establish a link between \mt hallucinations under domain shift and exposure bias by showing that Minimum Risk Training, a training objective which addresses exposure bias, can reduce the frequency of hallucinations. However, these insights do not yet provide practical strategies for handling \mt hallucinations.

\looseness=-1
A complementary approach to diagnosing hallucinations is to identify their symptoms via model introspection at inference time. However, there lacks a systematic study of hallucinations from the model's internal perspective. Previous works are either limited to an interpretation method that is tied to an outdated model architecture~\cite{LeeFiratAgarwalFannjiangSussillo2018} or to pseudo-hallucinations~\cite{Voita2021}.
In this paper, we propose to shed light on the decoding behavior of hallucinations on both artificially perturbed and natural inputs through model introspection based on Layerwise Relevance Propagation~(\lrp)~\cite{Bach2015}, which is applicable to a wide range of neural model architectures. We focus on \mt tasks with the widely used Transformer model~\cite{Vaswani2017}, and examine existing and new hypotheses for how hallucinations are produced. These hypotheses share the intuition that anomalous patterns of contributions from source tokens are indicative of hallucinations, but operationalize it differently.

The \textbf{\LowSrcContribHypo} introduced by \citet{Voita2021} states that hallucinations occur when \nmt overly relies on the target context over the source. They test the hypothesis by inspecting the relative source and target contributions to \nmt predictions on Transformer models using \lrp. However, their study is limited to pseudo-hallucinations produced by force decoding with random target prefixes. This work will test this hypothesis on actual hallucinations generated by \nmt models.

The \textbf{\LocalSrcContribHypo} introduced by \citet{LeeFiratAgarwalFannjiangSussillo2018} states that hallucinations occur when \nmt model overly relies on a small subset of source tokens across all generation steps. They test it by visualizing the dot-product attention in RNN models, but it is unclear whether these findings generalize to other model architectures. In addition, they only study hallucinations caused by random token insertion. This work will test this hypothesis on hallucinations under various types of source perturbations as well as on natural inputs, and will rely on \lrp to quantify token contributions more precisely than with attention.

\looseness=-1
Inspired by the previous observation on attention matrices that an \nmt model attends repeatedly to the same source tokens throughout inference when it hallucinates~\cite{LeeFiratAgarwalFannjiangSussillo2018,Berard2019Naver} or generates a low-quality translation~\cite{RiktersF2017}, we formalize this observation as the \textbf{\StaticSrcContribHypo} \---\ the distribution of source contributions remains static along inference steps when an \nmt model hallucinates. 
While prior work \cite{LeeFiratAgarwalFannjiangSussillo2018,Berard2019Naver,RiktersF2017} focuses on the static attention to the EOS or full-stop tokens, this hypothesis is agnostic about which source tokens contribute. 
Unlike the \LowSrcContribHypo, this hypothesis exclusively relies on the source and does not make any assumption about relative source versus target contributions. 
Unlike the \LocalSrcContribHypo, this hypothesis is agnostic to the proportion of source tokens contributing to a translation.

In this work, we evaluate in a controlled fashion how well each hypothesis explains detached hallucinations, first on artificially perturbed samples that let us contrast hallucinated vs. non-hallucinated outputs in controlled settings~(\S~\ref{sec:exp_perturbed_hallucinations}) and second, on natural source inputs that let us test the generalizability of these hypotheses when they are used to automatically detect hallucinations in more realistic settings~(\S~\ref{sec:exp_natural_hallucinations}).\footnote{Code and data are released at \url{https://github.com/weijia-xu/hallucinations-in-nmt}}
\section{Study of Hallucinations under Perturbations via Model Introspection}
\label{sec:exp_perturbed_hallucinations}

\looseness=-1
Hallucinations are typically rare and difficult to identify in natural datasets. To test the aforementioned hypotheses at scale, we first exploit the fact that source perturbations exacerbate \nmt hallucinations~\cite{LeeFiratAgarwalFannjiangSussillo2018,RaunakMenezesJunczys-Dowmunt2021}.  We construct a perturbation-based counterfactual hallucination dataset on English$\rightarrow$Chinese by automatically identifying hallucinated \nmt translations given perturbed source inputs and contrast them with the \nmt translations of the original source~(\S~\ref{subsec:source_perturbations}). This dataset lets us directly test the three hypotheses by computing the relative token contributions to the model's predictions using \lrp~(\S~\ref{subsec:lrp}), and conduct a controlled comparison of patterns on the original and hallucinated samples~(\S~\ref{subsec:results_perturbed_hallucinations}).

\subsection{Perturbation-based Hallucination Data}
\label{subsec:source_perturbations}

To construct the dataset, we randomly select~$50k$ seed sentence pairs to perturb from the \nmt training corpora, and then we apply the following perturbations on the source sentences:\footnote{For better contrastive analysis, we select samples with source length of~$n=30$ and clip the output length by~$T=15$.}

\begin{itemize}
    \item We randomly misspell words by deleting characters with a probability of~$0.1$, as~\citet{KarpukhinLEG2019} show that a few misspellings can lead to egregious errors in the output.
    \item We randomly title-case words with a probability of~$0.1$, as~\citet{BerardCDRMN2019} find  that this often leads to severe output errors.
    \item We insert a random token at the beginning of the source sentence, as  \citet{LeeFiratAgarwalFannjiangSussillo2018,RaunakMenezesJunczys-Dowmunt2021} find it a reliable trigger of hallucinations. The inserted token is chosen from $100$ most frequent, $100$ least frequent, mid-frequency tokens~(randomly sampled~$100$ tokens from the remaining tokens), and punctuations.
\end{itemize}

Inspired by \citet{LeeFiratAgarwalFannjiangSussillo2018}, we then identify hallucinations using heuristics that compare the translations from the original and perturbed sources.
We select samples whose original \nmt translations~$y'$ are of reasonable quality compared to the reference~$y$~(i.e. $bleu(y, y') > 0.3$).
The translation of a perturbed source sentence~$\tilde{y}$ is identified as a hallucination if it is very different from the translation of the original source~(i.e. $bleu(y', \tilde{y}) < 0.03$) and is not a copy of the perturbed source~$\tilde{x}$~(i.e. $bleu(\tilde{x}, \tilde{y}) < 0.5$).\footnote{The \bleu thresholds are selected based on manual inspection of the translation outputs.}
This results in~$623$,~$270$, and~$1307$ contrastive pairs of the original~(non-hallucinated) and hallucinated translations under misspelling, title-casing, and insertion perturbations, respectively.

We further divide the contrastive pairs into degenerated and non-degenerated hallucinations. Degenerated hallucinations are ``bland,
incoherent, or get stuck in repetitive loops''~\cite{HoltzmanBFC2019}, i.e. hallucinated translations that contain~$3$ more repetitive $n$-grams than the source are identified as degenerated hallucinations, while the non-degenerated group contains relatively fluent but hallucinated translations. 

\subsection{Measuring Relative Token Contributions}
\label{subsec:lrp}

We test the three source contribution hypotheses described in \S~\ref{sec:hypotheses} on the resulting dataset by contrasting the contributions of relevant tokens to the generation of a hallucinated versus a non-hallucinated translation using \lrp~\cite{Bach2015}. 
\lrp decomposes the prediction of a neural model computed over an input instance into relevance scores for input dimensions. Specifically, \lrp decomposes a neural model into several layers of computation and measures the relative influence score~$R^{(l)}_i$ for input neuron~$i$ at layer~$l$. 
Different from other interpretation methods that measure the absolute influence of each input dimension~\cite{AlvarezMelis2017,Ma2018,He2019}, \lrp adopts the principal that the relative influence~$R^{(l)}_i$ from all neurons at each layer should sum up to a constant:
\begin{equation}
    \sum_i R^{(1)}_i = \sum_i R^{(2)}_i = ... = \sum_i R^{(L)}_i = C
\end{equation}

To back-propagate the influence scores from the last layer to the first layer~(i.e. the input layer), we need to decompose the relevance score~$R^{(l+1)}_j$ of a neuron~$j$ at layer~$l+1$ into messages~$R^{(l, l+1)}_{i \leftarrow j}$ sent from the neuron~$j$ at layer~$l+1$ to each input neuron~$i$ at layer~$l$ under the following rules:
\begin{equation}
    R^{(l, l+1)}_{i \leftarrow j} = v_{ij} R^{(l+1)}_j, \, \sum_i v_{ij} = 1
\end{equation}
There exist several versions of \lrp, including \lrp-$\epsilon$, \lrp-$\alpha\beta$, and \lrp-$\gamma$, which compute~$v_{ij}$ differently~\cite{Bach2015,Binder2016,Montavon2019}. Following~\citet{Voita2021}, we use \lrp-$\alpha\beta$~\cite{Bach2015,Binder2016}, which defines~$v_{ij}$ such that the relevance scores are positive at each step. Consider first the simplest case of linear layers with non-linear activation functions:
\begin{equation}
    u^{(l+1)}_j = g(z_j),\, z_j = \sum_i z_{ij} + b_j,\, z_{ij} = w_{ij} u^{(l)}_{i}
\end{equation}
where~$u^{(l)}_{i}$ is the~$i$-th neuron at layer~$l$,~$w_{ij}$ is the weight connecting the neurons~$u^{(l)}_{i}$ and~$u^{(l+1)}_j$,~$b_j$ is a bias term, and~$g$ is a non-linear activation function. The~$\alpha\beta$ rule considers the positive and negative contributions separately:
$$
    z^+_{ij} = \max(z_{ij}, 0),\, b^+_j = \max(b_j, 0)
$$
$$
    z^-_{ij} = \min(z_{ij}, 0),\, b^-_j = \min(b_j, 0)
$$
and defines~$v_{ij}$ by the following equation:
\begin{equation}
    v_{ij} = \alpha \cdot \frac{z^+_{ij}}{\sum_i z^+_{ij} + b^+_j} + \beta \cdot \frac{z^-_{ij}}{\sum_i z^-_{ij} + b^-_j}
\end{equation}
Following~\citet{Voita2021}, we use~$\alpha=1,\, \beta=0$. 
This rule is directly applicable to linear, convolutional, maxpooling, and feed-forward layers. To back-propagate relevance scores through attention layers in the Transformer encoder-decoder model~\cite{Vaswani2017}, we follow the propagation rules in~\citet{Voita2021}, where the weighting~$v_{ij}$ is obtained by performing a first order Taylor expansion of each neuron~$u^{(l+1)}_j$.

In the context of \nmt, \lrp ensures that, at each generation step~$t$, the sum of contributions~$R_t(x_i)$ and~$R_t(y_j)$ from source tokens~$x_i$ and target prefix tokens~$y_j$ remains equal:
\begin{equation}
    \forall t, \sum_i R_t(x_i) + \sum_{j<t} R_t(y_j) = 1
\end{equation}
We further define normalized source contribution~$\bar{R}(x_i)$ at source position~$i$ averaged over all generation steps~$t$ as:
\begin{equation}
\label{eq:normalized_src_contribution}
    \bar{R}(x_i) = \frac{1}{T} \sum_t^T \frac{n \cdot R_t(x_i)}{ \sum_i^n R_t(x_i) }
\end{equation}
where~$n$ is the length of each source sequence and~$T$ is the length of the output sequence.

We then test the aforementioned hypotheses based on the distribution of relative token contributions and compare it with the attention matrix.

\subsection{\nmt Setup}
\label{sec:exp_setup}

We build strong Transformer models on two high-resource language pairs: English$\rightarrow$Chinese~(En-Zh) and German$\rightarrow$English~(De-En). They produce acceptable translation outputs overall, thus making hallucinations particularly misleading.

\paragraph{Data} For En-Zh, we use the $18M$ training samples from WMT18~\citep{BojarWMT18} and \textit{newsdev2017} as the validation set. For De-En, we use all training corpora from WMT21~\citep{WMT2021} except for ParaCrawl, which yields $5M$ sentence pairs after cleaning as in \citet{EdinburghWMT21}.\footnote{\url{https://github.com/browsermt/students/tree/master/train-student/clean}} We use \textit{newstest2019} for validation. 
We tokenize English and German sentences using the Moses scripts~\cite{Koehn2007Moses} and Chinese sentences using the Jieba segmenter.\footnote{\url{https://github.com/fxsjy/jieba}} For En-Zh, we train separate BPE models for English and Chinese using~$32k$ merging operations for each language. 
For De-En, we train a joint BPE model using~$32k$ merging operations.

\paragraph{Models}  All models are based on the \emph{base} Transformer~\citep{Vaswani2017}. 
We apply label smoothing of~$0.1$. We train all models using the Adam optimizer~\citep{KingmaB15} with initial learning rate of~$4.0$ and batch sizes of~$4,000$ tokens for maximum~$800k$ steps. 
We decode with beam search with a beam size of~$4$. The resulting \nmt models achieve close or higher \bleu scores than comparable published results.\footnote{The En-Zh model achieves~$33.5$ \bleu on \textit{newstest2017}, which is close to the $34.5$ achieved by the most comparable model in  \citet{XuC2018}. The De-En model achieves~$35.0$ \bleu on \textit{newstest2019}, which is higher than the strong baseline ~($29.6$ \bleu) from \citet{EdinburghWMT20}.}

\begin{figure}[t]
    \centering
    \includegraphics[width=0.3\textwidth,trim={2.5cm 0.5cm 1cm 0.5cm}]{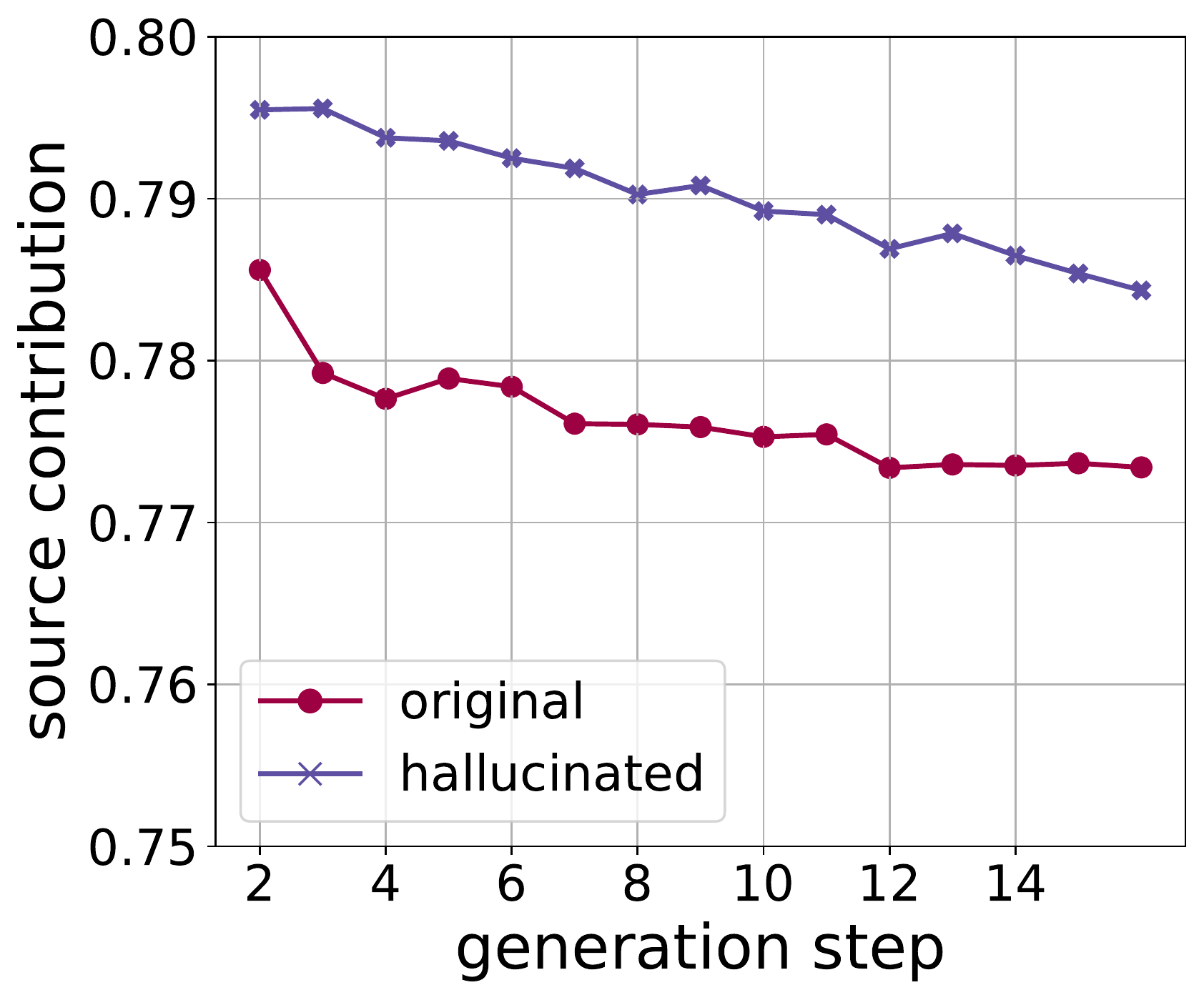}
\caption{Relative source contributions~$\sum_i R_t(x_i)$ at varying generation step~$t$ averaged over the original or hallucinated samples under a mixture of the misspelling, title-casing, and insertion perturbations.}
\label{fig:src_contrib_vs_tgt_pos}
\vspace{-10pt}
\end{figure}

\begin{table}[!t]
\centering
\scalebox{1}{
\begin{tabular}{lcccc}
\toprule
& \multicolumn{2}{c}{Contrib Ratio} & \multicolumn{2}{c}{Staticity} \\
& D & N & D & N \\
\midrule
Attention & -1.03$^\dagger$ & +0.51$^\dagger$ & 1.92$^\dagger$ & -1.10$^\dagger$ \\
\lrp & \textbf{-1.05$^\dagger$} & \textbf{-1.13$^\dagger$} & \textbf{3.16$^\dagger$} & \textbf{2.16$^\dagger$}\\
\midrule
\end{tabular}
}
\caption{Standardized mean difference in \HighContribRatio~(\textit{Contrib Ratio}) and \SrcContribStaticity~(\textit{Staticity})~(computed on attention and \lrp-based contribution matrices) between pairs of hallucinated and original samples. We show the score differences on degenerated~(\textit{D}) and non-degenerated~(\textit{N}) hallucinations separately. $\dagger$ indicates that the difference is statistically significant with~$p < 0.05$.}
\label{tab:perturbed_hallucination_results}
\vspace{-10pt}
\end{table}

\begin{figure*}[t]
    \begin{subfigure}[b]{0.28\textwidth}
        \includegraphics[width=\textwidth]{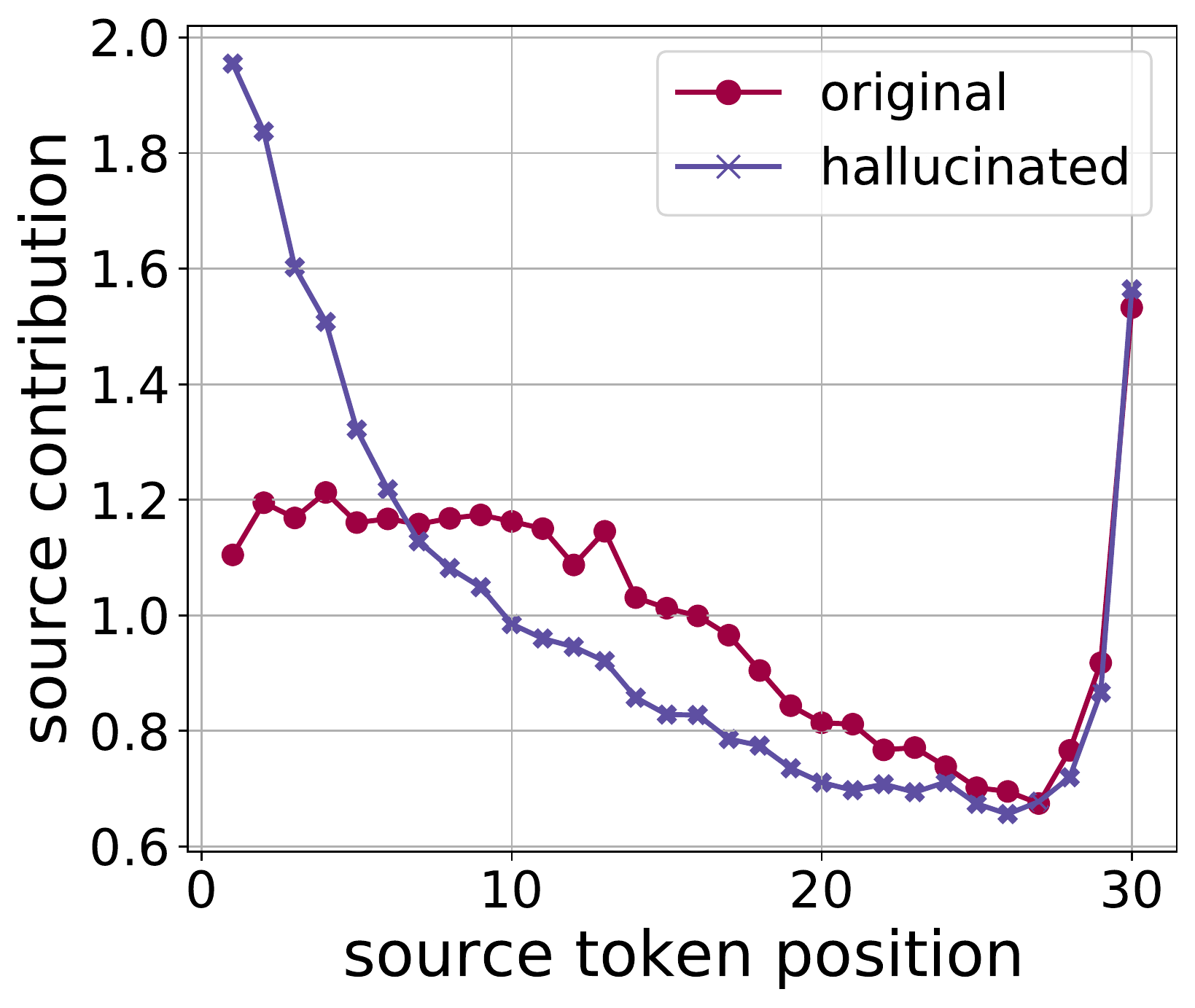}
        \caption{Misspelling}
    \end{subfigure}
	\hspace{1.5em}
    \begin{subfigure}[b]{0.28\textwidth}
        \includegraphics[width=\textwidth]{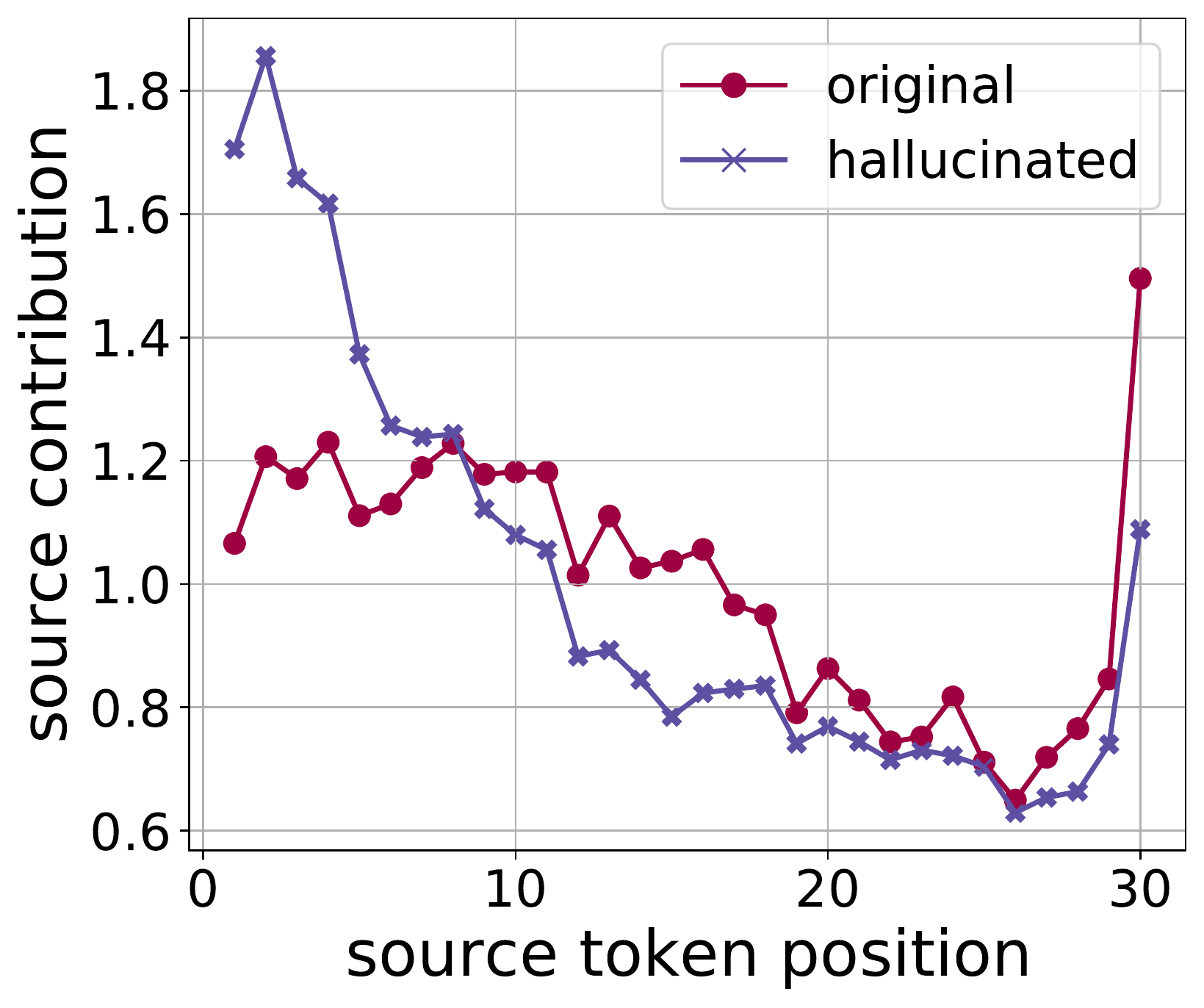}
        \caption{Title-Casing}
    \end{subfigure}
	\hspace{1.5em}
    \begin{subfigure}[b]{0.28\textwidth}
        \includegraphics[width=\textwidth]{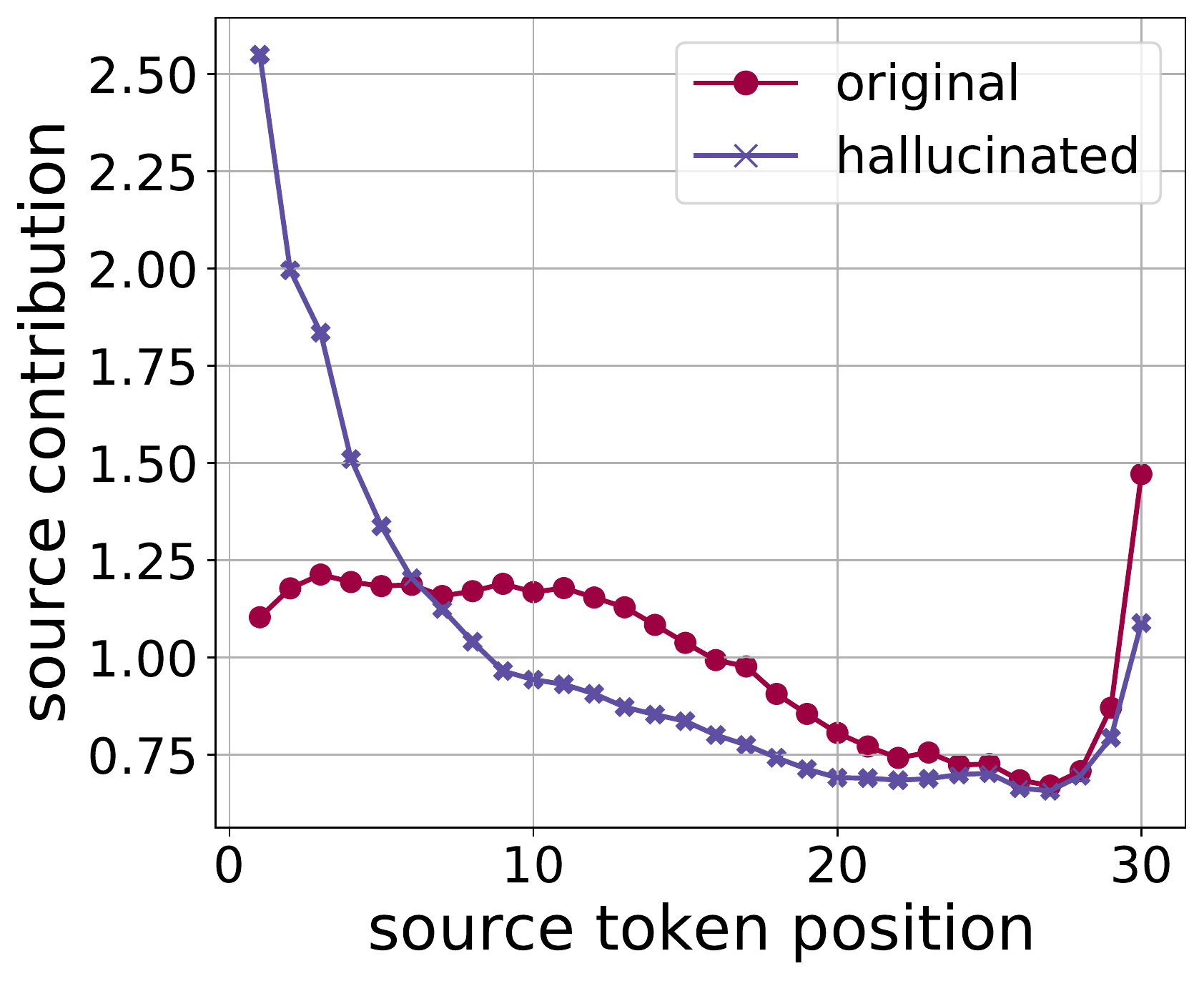}
        \caption{Insertion}
    \end{subfigure}
\caption{Normalized source contribution~$\bar{R}(x_i)$ (Eq.~\ref{eq:normalized_src_contribution}) at each source token position averaged over the original or hallucinated samples under~(a) misspelling,~(b) title-casing, and~(c) insertion perturbations.}
\label{fig:src_contrib_vs_src_pos}
\vspace{-10pt}
\end{figure*}

\begin{figure}[!t]
    \begin{subfigure}[b]{0.23\textwidth}
        \centering
        \includegraphics[width=\textwidth]{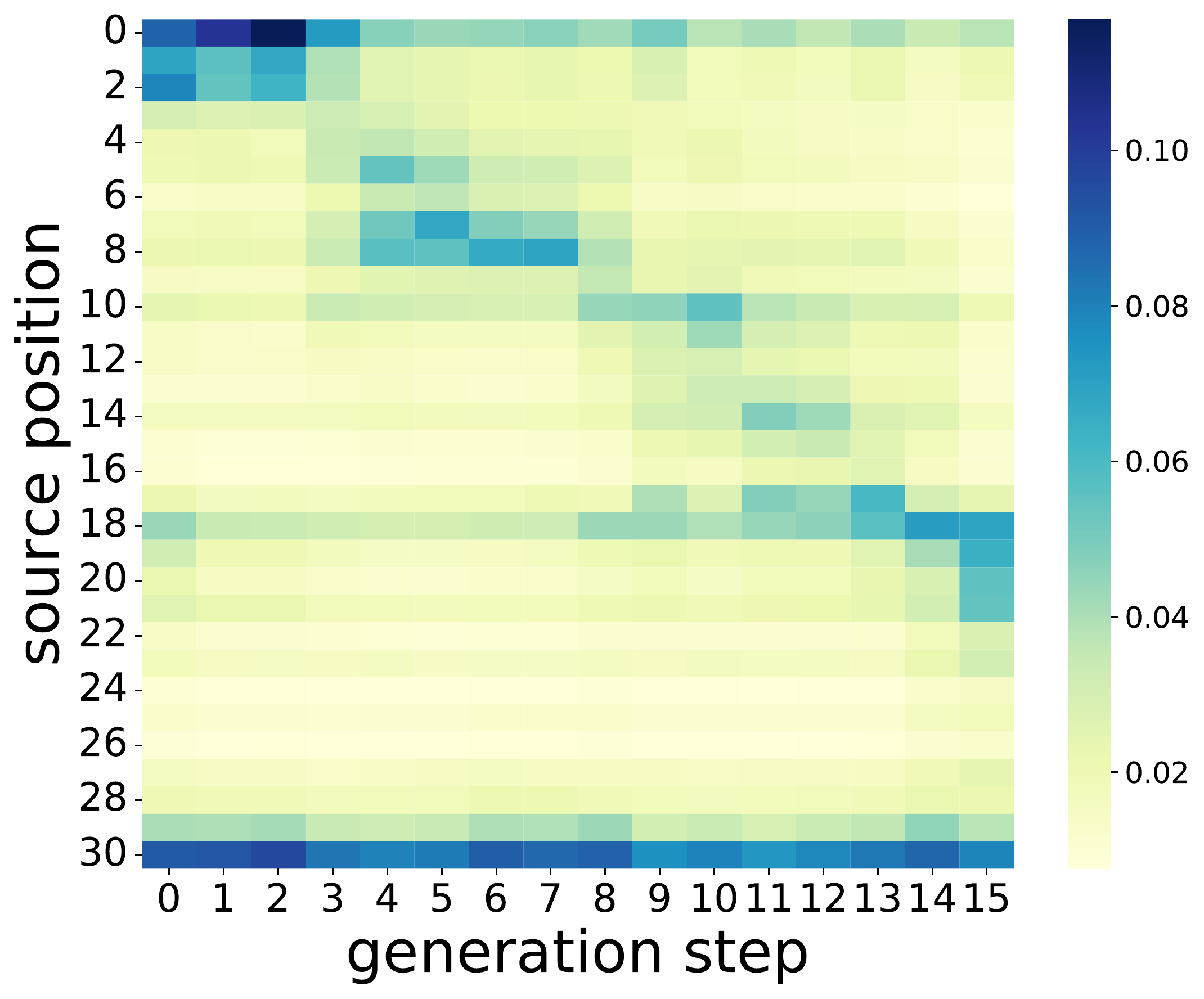}
        \caption{Original}
    \end{subfigure}
    \hfill
    \begin{subfigure}[b]{0.23\textwidth}
        \centering
        \includegraphics[width=\textwidth]{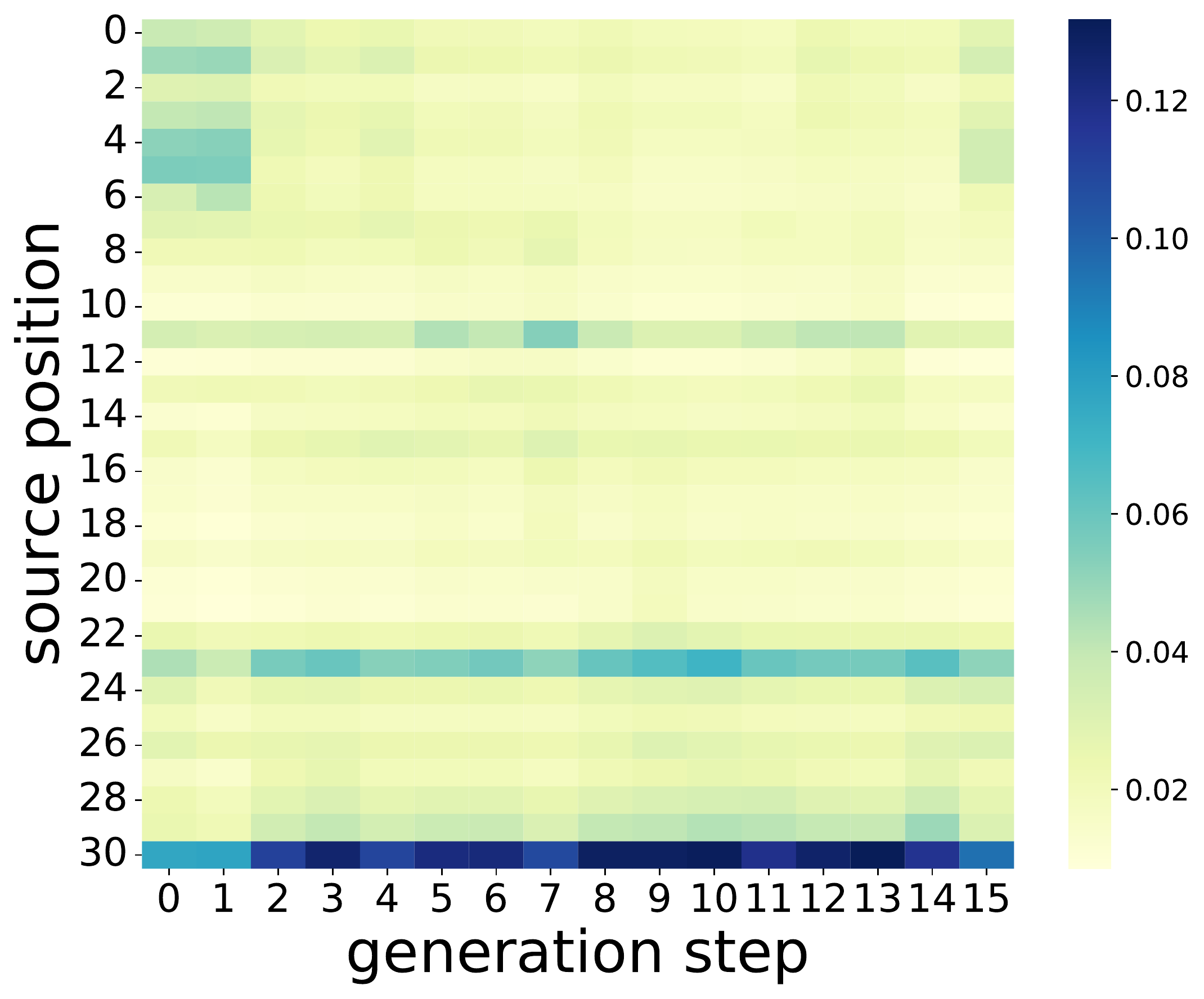}
        \caption{Hallucinated}
    \end{subfigure}
\caption{Heatmaps of relative contributions of source tokens~(y-axis) at each generation step~(x-axis) computed on the example of the original translation and the counterfactual hallucination from the perturbed source in Table~\ref{tab:example}. The source contribution distribution remains static across almost all generation steps on the hallucinated sample, unlike on the original sample.}
\label{fig:contrib_heatmap}
\vspace{-10pt}
\end{figure}

\subsection{Findings}
\label{subsec:results_perturbed_hallucinations}

We test the aforementioned hypotheses on the perturbation-based counterfactual hallucination dataset constructed on English$\rightarrow$Chinese.

First, we test the \textbf{\LowSrcContribHypo} by computing the relative source contributions~$\sum_{i=1}^n R_t(x_i)$ at each generation step~$t$, where~$n$ is the length of each source sequence.\footnote{Since \lrp ensures that the sum of source and target contributions at each generation step is a constant, we only visualize the relative source contributions.} We plot the average contributions over a set of samples in Figure~\ref{fig:src_contrib_vs_tgt_pos}. It shows that hallucinations under source perturbations have only slightly higher source contributions~($\Delta \approx 0.1$) than the original samples.  This departs from previous observations on pseudo-hallucinations \citep{Voita2021}, where the relative source contributions were lower on pseudo-hallucinations than on reference translations, perhaps because actual model outputs differ from pseudo-hallucinations created by inserting random target prefixes. We show that the hypothesis does not hold on actual hallucinations generated by the model itself. 

To explain this phenomenon, we further examine the source contribution from the end-of-sequence~(EOS) token. Previous works hypothesize that a translation is likely to be a hallucination when the attention distribution is concentrated on the source EOS token, which carries little information about the source~\cite{Berard2019Naver,RaunakMenezesJunczys-Dowmunt2021}. However, this hypothesis has only been supported by qualitative analysis on individual samples. Our quantitative results on the perturbation-based hallucination dataset do not support it, and align instead with the recent finding that the proportion of attention paid to the EOS token is not indicative of hallucinations \citep{GuerreiroVM2022}. Specifically, our results show that the proportion of source contribution from the EOS token is slightly higher on the original samples~($11.2$\%) than that on the hallucinated samples~($10.8$\%). We will show in the next part that the source contribution is more concentrated on the beginning than the end of the source sentence when the model hallucinates.

Second, we test the \textbf{\LocalSrcContribHypo} by computing the \textbf{\HighContribRatio~$r(\lambda_0)$} \---\ the ratio of source tokens with normalized contribution~$\bar{R}(x_i)$ larger than a threshold~$\lambda_0$:
\begin{equation}
\label{eq:high_contrib_ratio}
    r(\lambda_0) = \sum_{i=1}^n \mathbb{I}(\bar{R}(x_i) > \lambda_0) / n
\end{equation}
The ratio will be lower on hallucinated samples than on original samples if the hypothesis holds.  We compute the standardized mean difference in \HighContribRatio between the hallucinated and original samples (Table~\ref{tab:perturbed_hallucination_results}).\footnote{$\lambda_0$ is set to yield the largest score difference for each measurement type.} The negative score differences in \lrp-based scores supports the hypothesis, which is consistent with the findings of \citet{LeeFiratAgarwalFannjiangSussillo2018} based on attention weights. However, the attention-based score patterns are not consistent on degenerated and non-degenerated samples.

Furthermore, we investigate whether there is any positional bias for the local source contribution. We visualize the normalized source contribution~$\bar{R}(x_i)$ averaged over all samples with a source length of~$30$ in Figure~\ref{fig:src_contrib_vs_src_pos}. The source contribution of the hallucinated samples is disproportionately high at the beginning of a source sequence. 
By contrast, on the original samples, the normalized contribution is higher at the end of the source sequence, which could be a way for the model to decide when to finish generation. 
The positional bias exists not only on hallucinations under insertions at the beginning of the source, but also on hallucinations under misspelling and title-casing perturbations that are applied at random positions. 

\begin{figure}[!t]
    \centering
    \includegraphics[width=0.45\textwidth]{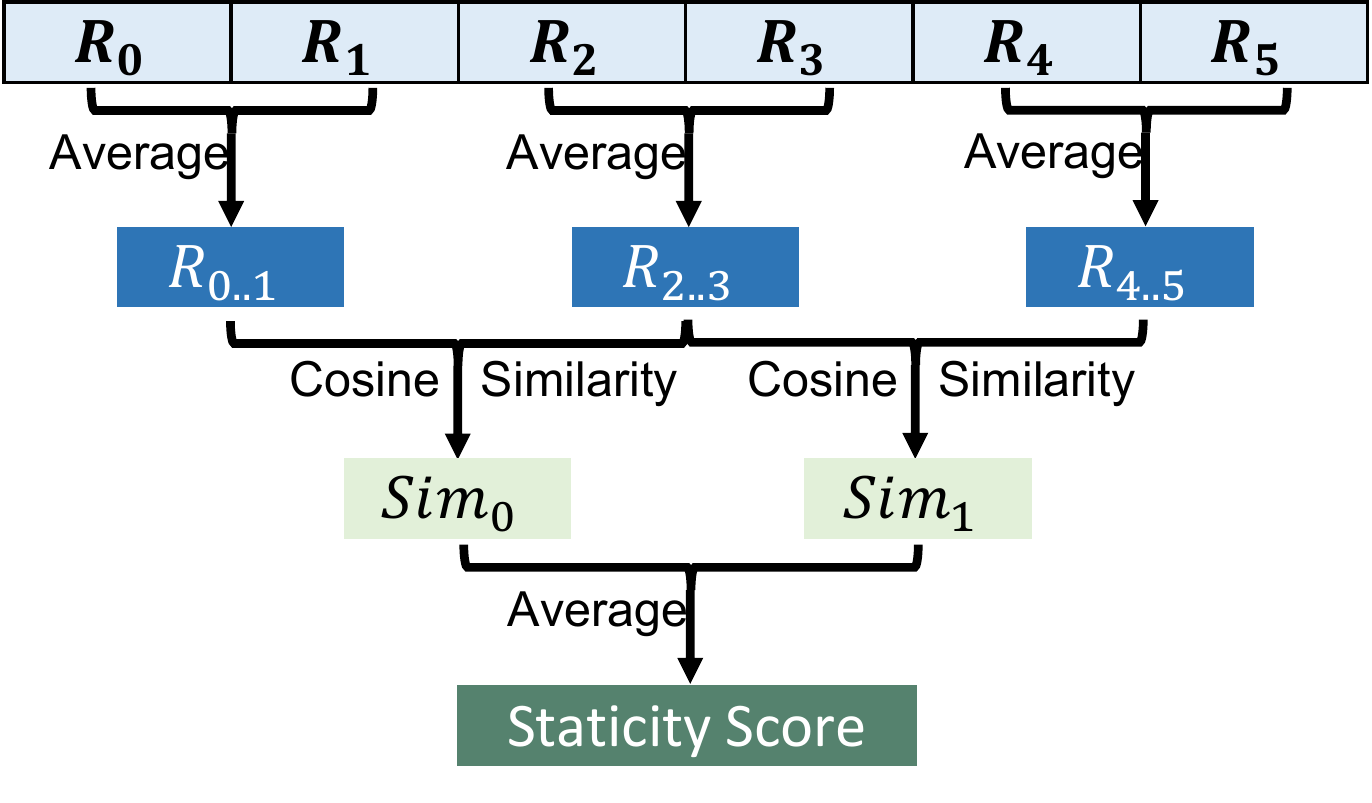}
\caption{Computing the \SrcContribStaticity of window size~$k=2$ given the source contribution vectors~$\boldsymbol{R_t} = [R_t(x_0) ... R_t(x_n)]$ at generation step~$t$.}
\label{fig:sliding_window}
\vspace{-10pt}
\end{figure}

Third, we examine the \textbf{\StaticSrcContribHypo} hypothesis by first visualizing the source contributions~$R_t(x_i)$ at varying source and generation positions on individual pairs of original and hallucinated samples. The heatmaps of source contributions for the example from Table~\ref{tab:example} are shown in Figure~\ref{fig:contrib_heatmap}. On the original outputs, the source contribution distribution in each column changes dynamically when moving horizontally along target generation steps. By contrast, when the model hallucinates, the source contribution distribution remains roughly static.

To quantify this pattern, we introduce \textbf{\SrcContribStaticity}, which measures how the source contribution distribution shifts over generation steps. Specifically, given a window size~$k$, we first divide the target sequence into several non-overlapping segments, each containing~$k$ tokens. Then, we compute the average vector over the contribution vectors~$\boldsymbol{R_t} = [R_t(x_0) ... R_t(x_n)]$ at steps~$t$ within each segment. Finally, we measure the cosine similarity between the average contribution vectors of adjacent segments and average over the cosine similarity scores at all positions as the final score~$s_k$ of window size~$k$. Figure~\ref{fig:sliding_window} illustrates this process for a window size of~$2$. 

Table~\ref{tab:perturbed_hallucination_results} shows the standardized mean difference in \SrcContribStaticity between the hallucinated and original samples in the degenerated and non-degenerated groups, taking the maximum staticity score among window sizes~$k \in [1, 3]$ for each sample. The positive differences in \lrp-based scores supports the \StaticSrcContribHypo \---\ the source contribution distribution is more static on the hallucinated samples than that on the original samples. Furthermore, \lrp distinguishes hallucinations from non-hallucinations better than attention, especially on non-degenerated samples where the translation outputs contain no repetitive loops.

In summary, we find that, when generating a hallucination under source perturbations, the \nmt model tends to rely on a small proportion of the source tokens, especially the tokens at the beginning of the source sentence. In addition, the distribution of the source contributions is more static on hallucinated translations than that on non-hallucinated translations. We turn to applying these insights on natural hallucinations next. 
\section{A Classifier to Detect Natural Hallucinations}
\label{sec:detection}

Based on these findings, we design features for a lightweight hallucination detector trained on samples automatically constructed by perturbations.

\paragraph{Classifier}
We build a small multi-layer perceptron~(\mlp) with a single hidden layer and the following input features:
\begin{itemize}
    \item \textbf{\NormSrcContrib} of the first~$K_1$ source tokens and the last~$K_1$ source tokens:~$\bar{R}(x_i)|i=1, ..., K_1, n-K_1+1, ..., n$~(where~$n$ is the length of the source sequence and~$K_1$ is a hyper-parameter), as we showed in the \LocalSrcContribHypo that the contributions of the beginning and end tokens distribute differently between hallucinated and non-hallucinated samples.
    \item \textbf{\SrcContribStaticity~$s_k$} given the source contributions~$R_t(x_i)$ and a window size~$k$ as defined in \S~\ref{subsec:results_perturbed_hallucinations}. We include the similarity scores of window sizes~$k = \{ 1, 2, ..., K_2 \}$ as input features, where~$K_2$ is a hyper-parameter.
\end{itemize}
This yields small classifiers with input dimension of $9$. For each language pair, we train~$20$ classifiers with different random seeds and select the model with the highest validation F1 score.

\paragraph{Data Generation} 
\looseness=-1
We construct the training and validation data using the same approach to constructing the perturbation-based hallucination dataset~(\S~\ref{subsec:source_perturbations}), but with longer seed pairs \---\ we randomly select seed sentence pairs with source length between~$20$ and~$60$ from the training corpora. We split the synthetic data randomly into the training~(around~$1k$ samples) and validation~(around~$200$ samples) sets with roughly equal number of positive and negative samples.
\section{Detecting Natural Hallucinations}
\label{sec:exp_natural_hallucinations}

While the hallucination classifier is trained on hallucinations from perturbations, we collect more realistic data to evaluate it against a wide range of relevant models. 

\subsection{Natural Hallucination Evaluation Set}
\label{sec:naturalhallucinations}

\begin{table}[!t]
\centering
\scalebox{1}{
\begin{tabular}{lrr}
\toprule
& En-Zh & De-En \\
\midrule
Detached hallucination & 111 & 189 \\
Non hallucination, including: \\
~~\textit{Faithful translation} & 154 & 153 \\
~~\textit{Incomplete translation} & 80 & 17 \\
~~\textit{Locally unfaithful} & 58 & 31 \\
~~\textit{Incomprehensible but aligned} & 5 & 33 \\
{\bf Total} & 408 & 423 \\
\hline
\end{tabular}
}
\caption{Human annotation label distribution on the En-Zh and De-En natural hallucination test sets (with random tie breaking on fine-grained labels; there are no ties on binary labels post-aggregation).}
\label{tab:eval_data}
\vspace{-10pt}
\end{table}

We build a test bed for detached hallucination detection for different language pairs and translation directions (En-Zh and De-En), and release the data together with the underlying \nmt models (described in \S~\ref{sec:exp_setup}). 

\looseness=-1
Since hallucinations are rare, we collect samples from large pools of out-of-domain data for our models to obtain enough positive examples of hallucinations for a meaningful test set. We use TED talk transcripts from the IWSLT15 training set~\cite{IWSLT15} for En-Zh, and the JRC-Acquis corpus~\cite{JRCAcquisData} of legislation from the European Union for De-En. To increase the chance of finding hallucinations, we select around~$200$,~$50$ and~$50$ translation outputs with low \bleu, low \comet~\cite{Rei2020}, or low \laser similarity~\cite{LASER2019} scores, respectively. We further combine them with~$50$ randomly selected samples.

Three bilingual annotators assess the faithfulness of the \nmt output given each input. 
While we ultimately need a binary annotation of outputs as hallucinated or not, annotators were asked to choose one of five labels to improve consistency:
\begin{itemize}
    \item {\it Detached hallucination}: a translation with large segments that are unrelated to the source.
    \item {\it Faithful translation}: a translation that is faithful to the source.
    \item {\it Incomplete translation}: a translation that is partially correct but misses part(s) of the source.
    \item {\it Locally unfaithful}: a translation that contains a few unfaithful phrases but is otherwise faithful. 
    \item {\it Incomprehensible but aligned}: a translation that is incomprehensible even though most phrases can be aligned to the source.
\end{itemize}
All labels except for the ``detached hallucination'' are aggregated into the ``non-hallucination'' category. The inter-annotator agreement on aggregated labels is substantial, with a Fleiss's Kappa~\cite{Fleiss1971} score of $FK=0.77$ for De-En and~$FK=0.64$ for En-Zh. Disagreement are resolved by majority voting for De-En, and by adjudication by a bilingual speaker for En-Zh. 
This yields~$27\%$ of detached hallucinations on En-Zh and~$45\%$ on De-En. The non-hallucinated \nmt outputs span all the fine-grained categories above, as can be seen in Table~\ref{tab:eval_data}.
Hallucinations are over-represented compared to what one might expect in the wild, but this is necessary to provide enough positive examples of hallucinations for evaluation.

\subsection{Experimental Conditions}

\subsubsection{Introspection-based Classifiers}

We implement the \textbf{\lrp-based classifier} described in \S~\ref{sec:detection}. To lower the cost of computing source contributions, 
we clip the source length at~$40$, and only consider the influence back-propagated through the most recent~$10$ target tokens \---\ prior work shows that nearby context is more influential than distant context~\cite{KhandelwalHQJ2018}. 
We tune the hyper-parameters~$K_1$ and~$K_2$ within the space~$K_1 \in \{1, 3, 5, 7, 9\},\, K_2 \in \{4, 8, 12, 16\}$ based on the average F1 accuracy on the validation set over three runs.
We compare it with an \textbf{attention-based classifier}, which uses the same features, but computes token contributions using attention weights averaged over all attention heads.

\subsubsection{Model-free Baselines}

We use three simple baselines to characterize the task.
The \textbf{random classifier} that predicts hallucination with a probability of~$0.5$. 
The \textbf{degeneration} detector marks as hallucinations 
degenerated outputs that contain~$K$ more repetitive $n$-grams than the source, where~$K$ is a hyper-parameter tuned on the  perturbation-based hallucination data.
The \textbf{\nmt probability scores} are used as a coarse model signal to detect hallucinations based on the heuristic that the model is less confident when producing a hallucination. The output is classified as a hallucination if the probability score is lower than a threshold tuned on the perturbation-based hallucination data. 

\subsubsection{Quality Estimation Classifier}

\looseness=-1
We also compare the introspection-based classifiers with a baseline classifier based on the state-of-the-art quality estimation model \---\ \textbf{\cometqe}~\cite{Rei2020COMETQE}. Given a source sentence and its \nmt translation, we compute the \cometqe score and classify the translation as a hallucination if the score is below a threshold tuned on the perturbation-based validation set.

\subsubsection{Large Pre-trained Classifiers}

We further compare the introspection-based classifiers with classifiers that rely on large pre-trained multilingual models, to compare the discriminative power of the source contribution patterns from the NMT model itself to extrinsic semantically-driven discrimination criteria. 

We use the cosine distance between the \textbf{\laser} representations~\cite{LASER2019,LASER2022} of the source and the \nmt translation. It classifies a translation as a hallucination if the distance score is higher than a threshold tuned on the perturbation-based validation set.

Inspired by local hallucination~\cite{ZhouNGDGZG2020} and cross-lingual semantic divergence~\cite{BriakouC2020} detection methods, we build an \textbf{\xlmr classifier} by fine-tuning the \xlmr model~\cite{Conneau2020XLMR} on synthetic hallucination samples. 
We randomly select~$50K$ seed pairs of source and reference sentences with source lengths between~$20$ and~$60$ from the parallel corpus and use the following perturbations to construct examples of detached hallucinations:
\begin{itemize}
    \item Map a source sentence to a random target from the parallel corpus to simulate natural, detached hallucinations.
    \item Repeat a random dependency subtree in the reference many times to simulate degenerated hallucinations.
    \item Drop a random clause from the source sentence to simulate natural, detached hallucinations.
\end{itemize}
We then collect diverse non-hallucinated samples: 
\begin{itemize}
    \item Original seed pairs provide faithful translations.
    \item Randomly drop a dependency subtree from a reference to simulate incomplete translations.
    \item Randomly substitute a phrase in the reference keeping the same part-of-speech to simulate translations with locally unfaithful phrases.
\end{itemize}
The final training and validation sets contain around~$300k$ and~$700$ samples, respectively.
We fine-tune the pre-trained model with a batch size of~$32$. We use the Adam optimizer~\cite{KingmaB15} with decoupled weight decay~\cite{Loshchilov2019} and an initial learning rate of~$2 \times 10^{-5}$. We fine-tune all models for~$5$ epochs and select the checkpoint with the highest F1 score on the validation set.

\begin{table*}[!t]
\centering
\scalebox{1}{
\begin{tabular}{l@{\hskip 0.2in}r@{\hskip 0.4in}rrrr@{\hskip 0.4in}rrrr}
\toprule
& \multirow{2}{*}{Params} & \multicolumn{4}{c}{De-En} & \multicolumn{4}{c}{En-Zh} \\
& & P & R & F1 & AUC & P & R & F1 & AUC \\
\midrule
\multicolumn{4}{l}{\textit{Model-free Baselines}} \\
Random & 0 & 44.0 & 49.9 & 46.8 & 50.2 & 27.6 & 49.8 & 35.5 & 48.0 \\
Degeneration & 1 & 49.1 & 59.3 & 53.7 & -- & 63.2 & 71.2 & 66.9 & -- \\
\nmt Score  & 1 & 33.3 & 3.4 & 6.2 & 37.7 & 35.4 & \textbf{91.9} & 51.1 & 49.8 \\
  \addlinespace[0.3cm]
\multicolumn{4}{l}{\textit{Quality Estimation Classifier}} \\
\cometqe  & $363M$ & 72.2 & 71.4 & \textbf{71.8} & 82.4 & 32.4 & \textbf{99.1} & 48.9 & 89.4 \\
 \addlinespace[0.3cm]
\multicolumn{4}{l}{\textit{Large Pre-trained Classifiers}} \\
\laser  & $45M$ & 81.6 & 54.0 & 65.0 & \textbf{89.5} & 54.6 & 64.0 & 58.9 & 75.3 \\
\xlmr  & $125M$ & 91.3 & 21.0 & 33.8 & 45.6 & \textbf{94.9} & \textbf{83.2} & \textbf{88.6} & \textbf{93.3} \\
 \addlinespace[0.3cm]
\multicolumn{4}{l}{\textit{Introspection-based Classifiers}} \\
Attention-based & $<400$ & 54.3 & \textbf{89.0} & 67.4 & 70.1 & 36.0 & 71.0 & 47.7 & 68.6 \\
\lrp-based & $<400$ & 87.3 & \textbf{76.2} & \textbf{81.2} & \textbf{91.4} & 87.5 & \textbf{85.6} & \textbf{86.4} & \textbf{96.5} \\
 \addlinespace[0.3cm]
\multicolumn{4}{l}{\textit{Ensemble Classifier}} \\
\lrp + \laser & $45M$ & \textbf{100.0} & 45.7 & 62.7 & -- & \textbf{94.5} & 59.5 & 72.9 & -- \\
\lrp + \xlmr  & $125M$ & 95.3 & 21.5 & 35.1 & -- & \textbf{97.6} & 72.4 & 83.1 & -- \\
\midrule
\end{tabular}
}
\caption{Precision (P), Recall (R), F1 and Area Under the Receiver Operating Characteristic Curve~(AUC) scores of each classifier on English-Chinese~(En-Zh) and German-English~(De-En) \nmt outputs~(means of three runs). We boldface the highest scores based on independent student’s t-test with Bonferroni Correction~($p < 0.05$). The \textit{Params} column indicates the total number of parameters used for each method (in addition to the \nmt parameters).}
\label{tab:main_results}
\vspace{-10pt}
\end{table*}

\subsection{Findings}

As shown in Table~\ref{tab:main_results}, we compare all classifiers against the baselines by the Precision, Recall and F1 scores. 
Since false positives and false negatives might have a different impact in practice (e.g., does the detector flag examples for review by humans, or entirely automatically? what is \mt used for?), we also report the Area Under the Receiver Operating Characteristic Curve~(AUC), which characterizes the discriminative power of each method at varying threshold settings. 

\paragraph{Main Results} The \lrp-based, \xlmr and the \laser classifiers are the best hallucination detectors, reaching AUC scores around 90s for either or both language pairs, which is considered outstanding discrimination ability \citep{Hosmer2013}.

\looseness=-1
The \lrp-based classifier is the best and most robust hallucination detector overall. It achieves higher F1 and AUC scores than \laser on both language pairs. Additionally, it outperforms \xlmr by +47 F1 and +46 AUC on De-En, while achieving competitive performance on En-Zh. This shows that the source contribution patterns identified on hallucinations under perturbations (\S~\ref{sec:exp_perturbed_hallucinations}) generalize as symptoms of natural hallucinations even under domain shift, as the domain gap between training and evaluation data is bigger on De-En than En-Zh.
It also confirms that \lrp provides a better signal to characterize token contributions than attention, improving F1 by~14-39 points and AUC by~21-28 points. These high scores represent large improvements of 41-54 points on AUC and 20-75 points on F1 over the model-free baselines. 

\paragraph{Model-free Baselines} These baselines shed light on the nature of the hallucinations in the dataset. The degeneration baseline is the best among them, with~$53.7$ F1 on De-En and~$66.9$ F1 on En-Zh, indicating that the Chinese hallucinations are more frequently degenerated than the English hallucinations from German. 
However, ignoring the remaining hallucinations is problematic, since they might be more fluent and thus more likely to mislead readers. The \nmt score is a poor predictor, scoring worse than the random baseline on De-En, in line with previous reports that \nmt scores do not capture faithfulness well during inference~\cite{WangTSL2020}. Manual inspection shows that the \nmt score can be low when the output is faithful  but contains rare words, and it can be high for a hallucinated output that contains mostly frequent words. 

\paragraph{Quality Estimation Classifier} 
The \cometqe classifier achieves higher AUC and F1 scores than the model-free classifiers, except for En-Zh, where the degeneration baseline obtains higher F1 than the \cometqe classifier. 
However, compared with the \lrp-based classifier, \cometqe lags behind by~9-38 points on F1 and~7-9 points on AUC.
This is consistent with previous findings that quality estimation models trained on data with insufficient negative samples~(e.g. \cometqe) are inadequate for detecting critical \mt errors such as hallucinations~\cite{Takahashi2021,Sudoh2021,GuerreiroVM2022}.

\paragraph{Pre-trained Classifiers} The performance of pre-trained classifiers varies greatly across language pairs. \laser achieves a competitive AUC score to the \lrp-based classifier on De-En but lags behind on En-Zh, perhaps because the \laser model is susceptible to the many rare tokens in the En-Zh evaluation data~(from TED subtitles).  \xlmr obtains better performance on En-Zh, approaching that of the \lrp-based classifier, but lags behind greatly on De-En. This suggests that the \xlmr classifier suffers from domain shift, which is bigger on De-En~(News$\rightarrow$Law) than En-Zh~(News$\rightarrow$TED). Fine-tuning the model on the synthetic training data generalizes more poorly across domains.
By contrast, the introspection-based classifiers are more robust.  

\paragraph{Ensemble Classifiers} 
The \laser and \xlmr classifiers emerge as the top classifiers apart from the \lrp-based one, but they make different errors than \lrp \---\ the confusion matrix comparing their predictions shows that the \laser and \lrp classifiers agree on~$68$-$78$\% of samples, while the \xlmr and \lrp classifiers agree on~$64$-$88$\% of samples. Thus an ensemble of \lrp + \laser or \lrp + \xlmr~(which detects hallucinations when the two classifiers both do so) yields a very high precision (at the expense of recall).

\begin{table}[]
\centering
\scalebox{1}{
\begin{tabular}{lcccc}
\toprule
& \multicolumn{2}{c}{De-En} & \multicolumn{2}{c}{En-Zh} \\
& F1 & AUC & F1 & AUC \\
\midrule
All features & \textbf{81.2} & \textbf{91.4} & \textbf{86.4} & \textbf{96.5} \\
- Src Contrib & 74.4 & \textbf{92.7} & \textbf{85.3} & \textbf{96.1} \\
- Staticity  & 50.7 & 76.6 & 58.3 & 80.0 \\
\midrule
\end{tabular}
}
\caption{Ablating the \NormSrcContrib~(\textit{Src Contrib}) and \SrcContribStaticity~(\textit{Staticity}) features used in the \lrp-based classifier. We boldface the highest scores based on independent student’s t-test with Bonferroni Correction~($p < 0.05$).}
\label{tab:ablation}
\vspace{-10pt}
\end{table}

\paragraph{\lrp Ablations} The \lrp-based classifier benefits the most from \SrcContribStaticity features (Table~\ref{tab:ablation}). Removing them hurts AUC by~$15$-$17$ points and F1 by~$28$-$31$, confirming that the \StaticSrcContribHypo holds on natural hallucinations. Ablating the \NormSrcContrib features also causes a significant drop in F1 on De-En, while its impact on En-Zh is not significant.

\begin{table}[t!]
\centering
\begin{tabular}{m{0.45\textwidth}}
\toprule
\textbf{Source:} C) DASS DIE WAREN IN DEM ZUSTAND IN DIE GEMEINSCHAFT VERSANDT WORDEN SIND, IN DEM SIE ZUR AUSSTELLUNG GESANDT WURDEN;\newline
\textbf{Correct Translation:} C) THAT THE GOODS WERE SHIPPED TO THE COMMUNITY IN THE CONDITION IN WHICH THEY ARE SENT FOR EXHIBITION;\newline
\textbf{Output:} C) THAT THE WOULD BE CONSIDERED IN THE COMMUNITY, IN which YOU WILL BE EXCLUSIVE;\\
\midrule
\end{tabular}
\caption{Example of a detached hallucination produced by the De-En \nmt being classified as non-hallucination by the \lrp-based classifier.}
\label{tab:lrp_failure_example}
\vspace{-10pt}
\end{table}

\paragraph{Error Analysis} 
\looseness=-1
Incomprehensible but aligned translations suffer from the highest false positive rate for the \lrp classifier, followed by incomplete translations. Additionally, the classifier can fail to detect hallucinations caused by the mistranslation of a large span of the source with rare or previously unseen tokens, rather than by pathological behavior at inference time as shown by the example in Table~\ref{tab:lrp_failure_example}.

\paragraph{Toward Practical Detectors} Detecting hallucinations in the wild is challenging since they tend to be rare and their frequency may vary greatly across test cases. 
We provide a first step in this direction by stress testing the top classifiers in an in-domain scenario where hallucinations are expected to be rare. Specifically, we randomly select~$10k$ English sentences from the \textit{``News Crawl: articles from 2021''} from WMT21~\citep{WMT2021} and use the En-Zh \nmt model to translate them into Chinese. We measure the \textit{Precision@20} for hallucination detection by manually examining the top-$20$ highest scoring hallucination predictions for each method. The \laser, \xlmr and \lrp-based classifiers evaluated above~(without fine-tuning in this setting) achieve~$35\%$,~$45\%$ and~$45\%$ Precision@20, respectively (compared to 0\% for the random baseline). More interestingly, after tuning threshold on the predicted probabilities~(which is originally set to~$0.5$) so that each classifier predicts hallucination~$1$\% of the time, the \lrp + \laser ensemble detects~$9$ hallucinations with a much higher precision of~$89\%$, and the \lrp + \xlmr ensemble detects~$12$ hallucinations with a precision of~$83\%$. These ensemble detectors thus have the potential to provide useful signals for detecting hallucinations even when they are needles in a haystack.

\subsection{Limitations} 
Our findings should be interpreted with several limitations in mind.
First, we exclusively study detached hallucinations in \mt. Thus, we do not elucidate the internal model symptoms that lead to partial hallucinations~\cite{ZhouNGDGZG2020}, although the methodology in this work could be used to shed light on this question.  Second, we work with \nmt models trained using the parallel data from \wmt without exploiting monolingual data or comparable corpora retrieved from collections of monolingual texts~(e.g. WikiMatrix~\cite{Schwenk2021WikiMatrix}). It remains to be seen whether hallucination symptoms generalize to \nmt models trained with more heterogeneous supervision.
Finally, we primarily test the hallucination classifiers in roughly balanced test sets, while hallucinations are expected to be rare in practice. We conducted a small stress test which shows the promise of our \lrp+\laser classifier in more realistic conditions. However, further work is needed to systematically evaluate how these classifiers can be used for hallucination detection in the wild. 

\section{Related Work}
\label{sec:related}

\looseness=-1
Hallucinations occur in all applications of neural models to language generation, including abstractive summarization~\cite{Falke2019,Maynez2020}, dialogue generation~\cite{Dusek2018NLGChallenge}, data-to-text generation~\cite{WisemanSR2017}, and machine translation~\cite{LeeFiratAgarwalFannjiangSussillo2018}. Most existing detection approaches view the generation model as a black-box, by ~1) training hallucination classifiers on synthetic data constructed by heuristics~\cite{ZhouNGDGZG2020,Santhanam2021}, or~2) using external models to measure the faithfulness of the outputs \---\ such as question answering or natural language inference models~\cite{Falke2019,DurmusHD2020}. These approaches ignore the signals from the generation model itself and could be highly biased by the heuristics used for synthetic data construction, or the biases in the external semantic models trained for other purposes. 
Concurrent to this work, \citet{GuerreiroVM2022} explore glass-box detection methods based on model confidence scores or attention patterns~(e.g. the proportion of attention paid to the EOS token and the proportion of source tokens with attention weights higher than a threshold). They evaluate these methods based on hallucination recall, and find that the model confidence is a better indicator of hallucinations than the attention patterns. In this paper, we investigated varying types of glass-box patterns based on the relative token contributions instead of attention, and find that these patterns yield more accurate hallucination detectors than model confidence.

Detecting hallucinations in \mt has not yet been directly addressed by the \mt quality estimation literature. Most quality estimation work has focused on predicting a direct assessment of translation quality, which does not distinguish adequacy and fluency errors~\cite{Guzman2019FLORES,Specia2020}. More recent task formulations target critical adequacy errors ~\cite{Specia2021}, but do not separate hallucinations from other error types, despite arguments that hallucinations should be considered separately from other \mt errors \cite{Shi2022}. The critical error detection task at \wmt2022 introduces an Additions error category, which refers to hallucinations where the translation content is only partially supported by the source~\cite{Zerva2022WMTQE}. Additions includes both detached hallucinations~(as in this work) and partial hallucinations. Methods for addressing all these tasks fall in two categories: ~1) black-box methods based on the source and output alone~\cite{Specia2009,KimLN2017,RanasingheOM2020}, and~2) glass-box methods based on features extracted from the \nmt model itself~\cite{RiktersF2017,YankovskayaTF2018,Fomicheva2020}. Black-box methods typically use resource-heavy deep neural networks trained on large amounts of annotated data. Our work is inspired by the glass-box methods that rely on model probabilities, uncertainty quantification, and the entropy of the attention distribution, but shows that relative token contributions computed through \lrp provide sharper features to characterize hallucinations.

\looseness=-1
This paper combines interpretability techniques to identify the symptoms of hallucinations. We adopt a saliency method to measure the importance of each input unit through a back-propagation pass~\cite{Simonyan2013,Bach2015,LiCHJ2016,DingXuKoehn2019a}.  While other saliency-based methods measure an abstract quantity reflecting the importance of each input feature by the partial derivative of the prediction with regard to each input unit~\cite{Simonyan2013}, \lrp~\cite{Bach2015} measures the proportional contribution of each input unit. This makes it well suited to compare model behavior across samples. Furthermore, \lrp does not require neural activations to be differentiable and smooth, and can be applied to a wide range of architectures, including RNN~\cite{DingLLS2017} and Transformer~\cite{Voita2021}. 
We apply this technique to analyze counterfactual hallucination samples inspired by perturbation methods~\cite{LiMJ2016,Feng2018,EbrahimiLD2018}, but crucially show that the insights generalize to natural hallucinations.

\section{Conclusion}

We contribute a thorough empirical study of the notorious but poorly understood hallucination phenomenon in \nmt, which shows that internal model symptoms exhibited during inference are strong indicators of hallucinations. Using counterfactual hallucinations triggered by perturbations, we show that distinctive source contribution patterns alone indicate hallucinations better than the relative contributions of the source and target. We further show that our findings can be used for detecting natural hallucinations much more accurately than model-free baselines and quality estimation models. Our detector also outperforms black-box classifiers based on pre-trained models. We release human-annotated test beds of natural English-Chinese and German-English hallucinations to enable further research.
This work opens a path toward detecting hallucinations in the wild and improving models to minimize hallucinations in \mt and other generation tasks.

\section*{Acknowledgements}
We thank our TACL Action Editor, the anonymous reviewers, and the UMD CLIP lab for their feedback. Thanks also to Yuxin Xiong for helping examine German outputs. This research is supported in part by an Amazon Machine Learning Research Award and by the National Science Foundation under Award No. 1750695. Any opinions, findings, conclusions or recommendations expressed in this material are those of the authors and do not necessarily reflect the views of the National Science Foundation.

\bibliography{anthology,reference}
\bibliographystyle{acl_natbib}

\end{document}